%% file: acl_latex.tex
\pdfoutput=1

\documentclass[11pt]{article}

\usepackage{acl}

\usepackage{times}
\usepackage{latexsym}

\usepackage[T1]{fontenc}

\usepackage[utf8]{inputenc}

\usepackage{microtype}

\usepackage{url}
\usepackage{inconsolata}
\usepackage{amssymb}
\usepackage{amsmath}
\usepackage{subfig}
\usepackage{graphicx}
\usepackage{graphics}
\usepackage{booktabs}
\usepackage{multirow}
\usepackage{color}
\usepackage{fixltx2e}
\usepackage{enumitem}
\usepackage{pgfplots}
\usepackage{makecell}
\usepackage{comment}
\usepackage{nicefrac}
\usepackage{amsfonts}
\usepackage{bm}
\usepackage{amsthm}

\usepackage{enumitem}
\setlist[itemize,enumerate]{leftmargin=*}
\usepackage[normalem]{ulem}
\usepackage{tikz}
\usepackage{xcolor}

\definecolor{fc}{HTML}{1E90FF}
\definecolor{h}{HTML}{228B22}
\definecolor{bias}{HTML}{87CEFA}
\definecolor{noise}{HTML}{8B008B}
\definecolor{conv}{HTML}{FFA500}
\definecolor{pool}{HTML}{B22222}
\definecolor{up}{HTML}{B22222}
\definecolor{view}{HTML}{FFFFFF}
\definecolor{bn}{HTML}{FFD700}
\tikzset{fc/.style={black,draw=black,fill=fc,rectangle,minimum height=1cm}}
\tikzset{h/.style={black,draw=black,fill=h,rectangle,minimum height=1cm}}
\tikzset{bias/.style={black,draw=black,fill=bias,rectangle,minimum height=1cm}}
\tikzset{noise/.style={black,draw=black,fill=noise,rectangle,minimum height=1cm}}
\tikzset{conv/.style={black,draw=black,fill=conv,rectangle,minimum height=1cm}}
\tikzset{pool/.style={black,draw=black,fill=pool,rectangle,minimum height=1cm}}
\tikzset{up/.style={black,draw=black,fill=up,rectangle,minimum height=1cm}}
\tikzset{view/.style={black,draw=black,fill=view,rectangle,minimum height=1cm}}
\tikzset{bn/.style={black,draw=black,fill=bn,rectangle,minimum height=1cm}}

\usepackage{arydshln}
\makeatletter
\def\adl@drawiv#1#2#3{%
        \hskip.5\tabcolsep
        \xleaders#3{#2.5\@tempdimb #1{1}#2.5\@tempdimb}%
                #2\z@ plus1fil minus1fil\relax
        \hskip.5\tabcolsep}
\newcommand{\cdashlinelr}[1]{%
  \noalign{\vskip 1.3pt
           \global\let\@dashdrawstore\adl@draw
           \global\let\adl@draw\adl@drawiv}
  \cdashline{#1}[.4pt/2pt]
  \noalign{\global\let\adl@draw\@dashdrawstore
           \vskip 1.3pt}}
\makeatother

\title{Scaling up \textsc{CometKiwi}:\\Unbabel-IST 2023 Submission for the Quality Estimation Shared Task}

\usepackage[misc]{ifsym}

\author{
     Ricardo Rei\thanks{~~Equal contribution. \Letter \, \url{ricardo.rei@unbabel.com}}\,\,$^{1,2,4}$,
     Nuno M. Guerreiro$^{*1,3,4}$,
     José Pombal$^{1}$,
     Daan van Stigt$^{1}$,
     \\
     \bf
     Marcos Treviso$^{3,4}$,
     Luisa Coheur$^{2,4}$,
     José G. C. de Souza$^{1}$,
     André F. T. Martins$^{1,3,4}$
     \\
     $^{1}$Unbabel, Lisbon, Portugal, \,\ $^{2}$INESC-ID, Lisbon, Portugal \\
     $^{3}$Instituto de Telecomunicações, Lisbon, Portugal \\
     $^{4}$Instituto Superior Técnico, University of Lisbon, Portugal
}

\begin{document}
\maketitle
\begin{abstract}
We present the joint contribution of Unbabel and Instituto Superior T\'ecnico to the WMT 2023 Shared Task on Quality Estimation (QE). Our team participated
on all tasks: sentence- and word-level quality prediction (task 1) and fine-grained error span detection (task 2). For all tasks, we build on the \textsc{CometKiwi-22} model \cite{rei-etal-2022-cometkiwi}. Our multilingual approaches are ranked first for all tasks, reaching state-of-the-art performance for quality estimation at word-, span- and sentence-level granularity. Compared to the previous state-of-the-art \textsc{CometKiwi-22} , we show large improvements in correlation with human judgements (up to 10 Spearman points). Moreover, we surpass the second-best multilingual submission to the shared-task with up to 3.8 absolute points. 
\end{abstract}

\section{Introduction}

Quality Estimation (QE) is the task of automatically assigning a quality score to a machine translation output without depending on reference translations \cite{specia2018quality}. This paper details the collaborative effort of Unbabel and Instituto Superior T\'ecnico (IST) in the WMT23 Quality Estimation shared task, which encompassed two primary tasks: (i) sentence- and word-level quality prediction and (ii) fine-grained error span detection.

As of last year, some language pairs in the test set were absent from the training data. To address this, following a similar approach to the previous year, our systems were developed to achieve good multilingual generalization and to accommodate previously unseen languages. To achieve this, we start by leveraging the direct assessments (DA) labeled data obtained from the WMT Metrics shared task from 2017 to 2020, the MLQE-PE dataset \cite{fomicheva2020mlqepe}, and the training data (DA) specifically annotated for Indian languages in the 2023 shared task edition. In total, these datasets encompass close to 1M annotations covering 38 language pairs. We start by constructing generic models using this corpus. These generic QE models were subsequently fine-tuned for this year's subtasks.

For Task 1 -- sentence-level, we fine-tuned our generic models exclusively with this year's DA data. The architecture of these models remains consistent with our submission from the previous year, but we employ XLM-R XL and XXL as pretrained encoders \cite{conneau-etal-2020-unsupervised}. For the word-level quality prediction task, we follow the successful approach of combining the sentence- and word-level signals into one loss during the finetuning step, which has yielded positive results in previous iterations \cite{rei-etal-2022-cometkiwi}. For fine-grained error span detection, we conducted experiments exploring various approaches that build upon our word-level and sentence-level strategies. In terms of contrasting systems, we explored UnbabelQi\footnote{\url{https://qi.unbabel.com/}} and GPT-4 \cite{openai2023gpt4}. For GPT-4, we used a prompt designed to predict both the location and severity of errors in each translation, akin to the approach used in AutoMQM \cite{Fernandes2023AutoMQM}.


Overall, our main contributions are: (i)~we introduce approaches for multilingual machine translation quality estimation that are consistently first-ranked at word-, span-, and sentence-level  granularity; (ii)~we explore different approaches to predict the span of problematic translations along with their error severities (\textsc{ok}, \textsc{Minor}, \textsc{Major}); (iii)~we publicly release two of our best models for research purposes (\textsc{CometKiwi} -XL\footnote{\url{https://huggingface.co/Unbabel/wmt23-cometkiwi-da-xl}} and -XXL\footnote{\url{https://huggingface.co/Unbabel/wmt23-cometkiwi-da-xxl}}). To the best of our knowledge, these are the largest open-source QE models publicly released. 

\textbf{Our submitted systems attain the top multilingual results in all tasks}: For Task 1 sentence-level prediction, our multilingual system achieves 59.4 Spearman correlation points, surpassing the second-best system by nearly 4 absolute points. For word-level, our system achieves a 31.7 MCC score, outperforming the second-best system by almost 2 absolute MCC points. 
For error span prediction, our multilingual system achieves a 22 F1.0 score, beating the second-best system by more than 5 $F_1$ points.




\section{Overview of the shared-task}

QE systems are designed according to the granularity in which predictions are made (e.g., sentence- or word-level QE). In sentence-level QE, the goal is to predict a single quality score $\hat{y} \in \mathbb{R}$ given the whole source and its translation as input. Word-level QE works at a lower granularity level, with the goal of predicting binary quality labels $\hat{y}_i \in \{\textsc{ok}, \textsc{bad}\}$ for all $1 \leq i \leq n$ \emph{machine-translated words}, indicating whether that word is a translation error.
In fine-grained error span detection, systems are tasked with flagging which parts of the segment, i.e., sequences of consecutive characters, contain errors. If an error span is found, the system has to point out its severity; in this shared task, an error span's severity can be classified as \textsc{minor} or \textsc{major}. We sometimes refer to the parts of the segment that do not belong to an error span as being labelled as \textsc{ok}. We participated on all tasks of this year's shared-task. We specify the language pairs and the released data below:

\paragraph{Task 1 -- Sentence-level quality prediction:} submissions were evaluated on correlation with DA annotations for 5 language pairs (\textit{en-mr}, \textit{en-hi}, \textit{en-ta}, \textit{en-te}, \textit{en-gu}) and MQM annotations for 3 language pairs (\textit{en-de}, \textit{zh-en} and \textit{he-en}). Training data was released or is available for all directions but \textit{he-en}.

\paragraph{Task 1 -- Word-level quality prediction:} submissions were evaluated on tags inferred from post-editions for 2 language pairs (\textit{en-fa}, \textit{en-mr}), and MQM annotations for 3 language pairs (\textit{en-de}, \textit{zh-en} and \textit{he-en}). No additional training or development data with word-level tags were made available. To the best of our knowledge, no word-level data is available for \textit{en-fa} and \textit{he-en}.

\paragraph{Task 2 -- Fine-grained error span detection:} submissions were evaluated on error spans obtained via MQM annotations for 3 language pairs (\textit{en-de}, \textit{zh-en} and \textit{he-en}). No training nor development data is available for \textit{he-en}.


\section{Implemented Systems} \label{sec:implemented_systems}


\input{figs/architecture}

We largely follow the architecture of \textsc{CometKiwi}~\cite{rei-etal-2022-cometkiwi} -- see Figure~\ref{fig:arch} for an illustration. 
We concatenate the machine translated sentence $\bm{t} = \langle t_1, ..., t_n \rangle$ and its source sentence counterpart $\bm{s} = \langle s_1, ..., s_m \rangle$ to serve as input to the encoder. This encoder then produces 
hidden state matrices 
$\bm{H}_0, ..., \bm{H}_{L}$ for each layer $0 \leq \ell \leq L$, where $\bm{H}_\ell \in \mathbb{R}^{(n+m)\times d}$, where $\ell=0$ corresponds to the embedding layer and $d$ is the hidden size. Following this, all hidden states are fed to a scalar mix module~\citep{peters-etal-2018-deep} that learns a weighted sum of the hidden states of each layer of the encoder, producing a sequence of aggregated hidden states $\bm{H}_{\mathrm{mix}}$ as follows:
\begin{equation}\label{eq:scalarmix}
    \bm{H}_{\mathrm{mix}} = \lambda \sum_{\ell = 0}^{L} \beta_{\ell} \bm{H}_{\ell}.
\end{equation}
Here $\lambda$ is a scalar trainable parameter, $\bm{\beta} \in \triangle^L$  is given by $\bm{\beta} = \textsf{sparsemax}(\bm{\phi})$ using a sparse transformation~\citep{martins2016softmax}, with $\bm{\phi} \in \mathbb{R}^L$ as learnable parameters, and where we denote by $\triangle^L := \{\bm{\beta} \in \mathbb{R}^L: \mathbf{1}^\top \bm{\beta} = 1, \bm{\beta} \ge 0\}$ the probability simplex.\footnote{As it has been shown in \cite{rei-etal-2022-searching} not all layers are relevant and thus, using \textsf{sparsemax} we learn to ignore layers that do not help in the task at hands.} 

For sentence-level models, we use the hidden state of the \texttt{<cls>} token as the sentence representation, which, in turn, is passed to a 2-layered feed-forward module in order to get a sentence score prediction $\hat{y} \in \mathbb{R}$. For word-level and error span detection models, we first retrieve the hidden state vectors associated with each each token in $\bm{t}$, and then pass them to a linear projection to get word-level predictions $\hat{y}_i \in \mathcal{Y}_{\textsc{wl}},\, \forall_{1 \leq i \leq n}$. The output space of the word-level predictions is different depending on whether the models are constructed for word-level quality prediction ($\mathcal{Y}_{\textsc{WL}} = \{\textsc{ok}, \textsc{bad}\}$), or error span detection ($\mathcal{Y}_{\textsc{WL}} = \{\textsc{ok}, \textsc{minor}, \textsc{major}\}$).

\paragraph{Pretrained multilingual encoders.} 
Similarly to~\citep{rei-etal-2022-cometkiwi}, we employ InfoXLM L~\citep{chi-etal-2021-infoxlm}.\footnote{\url{https://huggingface.co/microsoft/infoxlm-large}}  Additionally, we experiment with scaled-up multilingual encoders, including XLM-R XL,\footnote{\url{https://huggingface.co/facebook/xlm-roberta-xl}} 
and XLM-R XXL.\footnote{\url{https://huggingface.co/facebook/xlm-roberta-xxl}} 
InfoXLM L comprises 24 encoder blocks with 16 attention heads each, totaling 550M parameters. XLM-R XL and XLM-R XXL have 32 attention heads for each encoder block, 36 and 48 encoder blocks and a total of 3.5B and 10.7B parameters, respectively.

\paragraph{Generic models for all tasks.} We create, for each model size, a generic model that will then be further adapted to each separate task. To train these models, we use the collective corpora from 2017 to 2019 DA annotations of the WMT Translation shared task, and the MLQE-PE corpus~\cite{fomicheva2020mlqepe}. We include the human annotations respective to the language pairs of this year's shared task for 7 different language pairs: DA annotations for \textit{en-mr}, \textit{en-hi}, \textit{en-ta}, \textit{en-te}, \textit{en-gu}, and MQM annotations for \textit{en-de} and \textit{zh-en}. Overall, the generic models are trained on sentence-level quality prediction with over 940k samples with source, translation and quality score on 38 different language pairs. When presented with multiple DA scores for the same sentence pair, we used the z-score of the DAs for training but we first normalize the DAs between 0 and 1, where 1 represents a perfect translation and 0 a random one.

\paragraph{Task adaptation.} After having obtained the generic models, we will train models for each separate stream of the shared-task, i.e., sentence-level, word-level or error span prediction. To do so, we consider the multi-task optimization from~\citet{rei-etal-2022-cometkiwi} wherein sentence scores can be used alongside supervision from word-level tags. Formally,
\begin{align}
    \mathcal{L}_{\mathrm{\textsc{sl}}}(\theta) &= \frac{1}{2}(y - \hat{y}(\theta))^2 \\
    \mathcal{L}_{\mathrm{\textsc{wl}}}(\theta) &= -\frac{1}{n}\sum_{i=1}^{n} w_{y_i} \log p_\theta(y_i) \\
    \mathcal{L}(\theta) &= \lambda_{\mathrm{\textsc{sl}}} \mathcal{L}_{\mathrm{\textsc{sl}}}(\theta) + \lambda_{\mathrm{\textsc{wl}}} \mathcal{L}_{\mathrm{\textsc{wl}}}(\theta), \label{eq:comb_loss}
\end{align}
where $w \in \mathbb{R}^{|\mathcal{Y}_{\textsc{wl}}|}$ represents the class weights given for the word-level tags,\footnote{These parameters help control how much we penalize the different granularities of word-level errors.} and $\lambda_{\mathrm{\textsc{sl}}}, \lambda_{\mathrm{\textsc{wl}}} \in \mathbb{R}_+$ are used to weigh the sentence and word-level losses, respectively. Note that $\lambda_{\mathrm{\textsc{sl}}} = 1$ and $\lambda_{\mathrm{\textsc{wl}}} = 0$ yields a fully sentence-level model, whereas $\lambda_{\mathrm{\textsc{sl}}} = 0$ and $\lambda_{\mathrm{\textsc{wl}}} = 1$ yields a word-level model.

\paragraph{Using unconstrained models.} For error span detection, we evaluate UnbabelQi, an Unbabel demo QE system, alongside GPT4~\citep{openai2023gpt4}. We prompt GPT4 to produce an MQM annotation for each source-target pair, based on five-shot examples which vary
across language pairs but are consistent within segments of the same language pair. We also apply this system in Task 1, deriving a sentence-level score from error spans, in alignment with the MQM framework. This approach bears similarity to AutoMQM~\citep{Fernandes2023AutoMQM}.

\subsection{Task 1: Quality prediction}

After the pretraining phase, we \textit{further} \textbf{separately} adapt the generic models to the released DA and MQM data for this year's shared task.

\subsubsection{Sentence-level quality prediction}\label{subsubsec:sentence-level}

\paragraph{Adaptation for sentence-level.} To further adapt the models to this year's language pairs, we fine-tuned the generic models using, exclusively, the newly released DA annotations from this year. This approach yields additional improvements for those languages. In the case of the MQM language pairs, our preliminary experiments revealed that attempting significant performance improvements on the MQM data led to noteworthy drops in correlations for the other language pairs using DAs. Consequently, for the MQM language pairs, we opted to employ the generic models as they are.


\paragraph{Ensembling models.} Similarly to~\citet{rei-etal-2022-cometkiwi}, we use Optuna \cite{optuna_2019} to assemble four models -- two XL and two XXL -- into a single system. We do so by finding the optimal weights for each language pair among these four multilingual models, and combining their predictions according to those weights. Notably, the XXL models are generic models, whereas the two XL checkpoints were further optimized with this year's shared task data. As expected, the XL models carry more weight for Indian languages, while the XXL generic models were deemed more crucial for MQM languages.





\begin{table*}[t]
    \centering
    \small    
    \begin{tabular}{l ccccc c@{} cccc}
        \toprule
        & \multicolumn{5}{c}{\bf DA} & & \multicolumn{3}{c}{\bf MQM} \\
        \cmidrule{2-6}
        \cmidrule{8-10}
        \bf Encoder & \bf en-mr & \bf en-hi & \bf en-ta & \bf en-te &\bf en-gu & & \bf en-de & \bf zh-en & \bf he-en$^\dagger$ & \bf avg. \\ \midrule
        2nd place (TBA) & & & & & & & & & & 0.556 \\

        \multicolumn{11}{c}{\textit{CometKiwi-22~\citep{rei-etal-2022-cometkiwi}}} \\
        InfoXLM L & 0.625 & 0.394 & 0.549 & 0.229 & 0.577 & & 0.413 & 0.476  & 0.619 & 0.485 \\\midrule
        \multicolumn{11}{c}{\textit{Generic models}} \\
        InfoXLM L & 0.661 & 0.505 & 0.641 & 0.282 & 0.661 & & 0.422 & 0.448  & 0.610 & 0.529 \\
        XLM-R XL & 0.664 & 0.536 & 0.607 & 0.335 & 0.637 & & 0.422 & 0.469  & 0.624 & 0.537 \\
        XLM-R XXL & 0.685 & 0.520 & 0.670 & 0.326 & 0.655 & & 0.443 & 0.476 & 0.662 & 0.555\\ \midrule
        \multicolumn{11}{c}{\textit{Further adapted models for sentence-level}} \\
        XLM-R XL & 0.684 & 0.583 & 0.682 & 0.386 & 0.683 & & 0.434 & 0.441 & \textbf{0.696} & 0.574\\
        XLM-R XXL & 0.693 & 0.555 & 0.738 & 0.359 & 0.701 & & 0.434 & 0.457 & 0.661 & 0.575\\\midrule
        \multicolumn{11}{c}{\textit{Final Ensemble}}\\
        Ensemble 4x & \bf 0.702 & \bf 0.598 & \bf 0.739 &  \bf 0.389 & \bf 0.714 & &\bf 0.448 &\bf 0.493  & 0.668 & \bf 0.594 \\ \midrule\midrule
        \multicolumn{11}{c}{\textit{GPT4-based model}}\\
        GPT4-QE & 0.379 & 0.212 & 0.146 & 0.174 & 0.297 & & 0.442 & 0.412 & 0.488 & 0.319\\
        \bottomrule
    \end{tabular}
    \caption{Results for sentence-level QE in terms of Spearman correlation. We represent zero-shot LPs with $\dagger$.
    }
    \label{tab:results_task_1_DA}
\end{table*}

\subsubsection{Word-level quality prediction}
\label{subsubsec:word_level}
For the word-level QE tasks, we experimented with both the multi-task setting and word-labels only.

\paragraph{Training word-level models.} This year, no training or development data with word-level tags were made available. As such, the training data for our models consists of the training data used in \citet{rei-etal-2022-cometkiwi}, combined with the development sets from the 2022 WMT Shared Task. As the word-level task was going to be tested in a zero-shot scenario for two out of five language pairs (\textit{en-fa}, \textit{he-en}), contrary to \citet{rei-etal-2022-cometkiwi}, we do not prepend a language prefix to the beginning of the source and target segments during training. Moreover, for the post-edit (PE) models, we removed samples from two language pairs (\textit{ps-en} and \textit{en-cs}) from the training data. We did so to assess, during validation, the models' capability to generalise in a zero-shot scenario. For the MQM models, we used all available annotations, including those in \textit{en-ru}. 

\paragraph{Ensembling models.} For word-level we followed a similar ensembling technique used for sentence-level. Specifically, we combined multiple systems trained with different hyperparameters, encoder size and pre-training setups. In the case of word-level predictions, we aggregate multiple predictions into OK/BAD tags by following the \textit{ensemble-tags} procedure from \citet{rei-etal-2022-cometkiwi}. In this approach, we combine the predicted tags of each model: for every input segment, we get a combined tag, $\alpha \sum_{i \in \mathcal{M}} w_i c_i$, where $c_i$ is the tag predicted by the model and $\alpha$ is the weight for the \textsc{bad} tag. We use Optuna to determine the optimal weights $w_i$ for each model and the optimal \textsc{bad} weight $\alpha$ for each LP. In the final submission, we combine six models (five PE models and one MQM model). Five of these models use InfoXLM as the encoder model, and one PE model uses XLM-R XL.\footnote{We found it hard to obtain performance boosts by scaling up to XLM-R XL on the word-level task. As such, we did not experiment with XLM-R XXL.} Refer to Table \ref{tab:results_task_1_word_level_da} for the test set results.

\subsection{Task 2: Fine-grained error span detection}\label{subsubsec:error-span}
In this task, we investigated three distinct approaches. The first approach extends word-level models by modifying their output predictions. More precisely, it involves transforming consecutively predicted \textsc{bad} tags into character-level error spans, rather than categorizing individual words based on the first subword. To determine the error severities of these spans, we considered two options: labeling all the subwords within the span as either \textsc{minor} or \textsc{major}. Our best results were achieved with the latter approach.


%

The second approach leverages \textsc{xComet}~(TBA)\footnote{Further details about \textsc{xComet} will be provided soon.} 
in conjunction with a pseudo-reference obtained from DeepL or Google Translate.\footnote{We choose the best translation using the generic XXL model from task 1.} Similar to our models from Task 1 word-level, \textsc{xComet} is trained with a multitask objective. Additionally, \textsc{xComet} is simultaneously optimized for both reference-free and reference-based evaluation, following \textsc{UniTE} \cite{wan-etal-2022-unite}. During inference, \textsc{xComet} can leverage a reference translation to enhance error identification. Since we employ a pseudo-reference that may contain translation errors, we initially assess the quality of the pseudo-reference using a generic QE system from Task 1 (reference\_score). For all pseudo-references with a score below $0.5$, we run \textsc{xComet} with QE-only input. For pseudo-references scoring above $0.5$, the input weights for \textsc{xComet} are determined as follows:
{
\begin{align*}
\text{diff} &= 1 - \text{reference\_score} \\
\text{src\_weight} &= 2 \cdot \text{diff} \\
\text{ref\_weight} &= (1 - \text{src\_weight}) \cdot 0.4 \\
\text{uni\_weight} &= (1 - \text{src\_weight}) \cdot 0.6
\end{align*}
}%

Here, \texttt{src\_weight} represents the weight assigned to the source-only input, \texttt{ref\_weight} denotes the typical metric input (reference-only input), and \texttt{uni\_weight} represents a unified input where the model receives all three sentences (translation, source, and reference). Notably, for pseudo-references with a QE score of 1, we rely solely on a reference-only input and the unified input. We refer to this approach as x\textsc{comet-ps-ref}.

We also contrast the aforementioned approaches with two unconstrained QE systems: UnbabelQi and GPT-4, as mentioned in Section~\ref{sec:implemented_systems}. We refer to these approaches as \textsc{UnbabelQi} and \textsc{GPT4-QE}, respectively.

\begin{table*}[t]
    \centering
    \small    
    \begin{tabular}{l cc c@{} cccc}
        \toprule
        & \multicolumn{2}{c}{\bf Post-edit} & & \multicolumn{3}{c}{\bf MQM} \\
        \cmidrule{2-3}
        \cmidrule{5-7}
        \bf Method & \bf en-fa$^\dagger$ & \bf en-mr & & \bf en-de & \bf zh-en & \bf he-en$^\dagger$ & \bf avg. \\
        \midrule
        Baseline~\citep{rei-etal-2022-cometkiwi} & 0.293 & 0.287 & & 0.179 &	0.225  & 0.275 & 0.226 \\
        2nd place (TBA) & & & & & & & 0.298 \\
        \midrule
        \multicolumn{8}{c}{\textit{Adapted models for word-level}} \\
        PE model (InfoXLM L) & 0.343 &	0.343 & & 0.227 &	0.253  & 0.382 & 0.310 \\
        PE model (XLM-R XL) & 0.325 &	0.344 & & \bf 0.255 &	0.197  & 0.306 & 0.285 \\\cdashlinelr{1-8}
        MQM model (InfoXLM L) & 0.296 & 0.252 & & 0.215 &	0.269  & 0.334 & 0.273 \\
        \midrule
        \multicolumn{8}{c}{\textit{Final Ensemble}} \\
        Ensemble PE + MQM & \bf 0.345 & \bf 0.347 & & 0.246 &	\bf 0.302  & \bf 0.402 & \bf 0.317 \\
        \bottomrule
    \end{tabular}
    \caption{Results for word-level QE in terms of MCC for the post-edit and MQM LPs. The ensemble is composed by multiple post-edit and MQM models. We represent zero-shot LPs with $\dagger$.
    }
    \label{tab:results_task_1_word_level_da}
\end{table*}

\section{Experimental Results} 
\label{sec:experiments}
We present the results on the official test set for each of the tasks for multiple model/data configurations. Sentence-level submissions were evaluated using the Spearman rank correlation. Pearson and Kendall correlation were also used as secondary metrics, but here we report only Spearman since it was the primary metric used to rank systems. word-level submission were evaluated using MCC, $F_1$-OK, and $F_1$-BAD, but we report only MCC as it was considered the main metric.
Error span detection was evaluated using $F_1$ score in which the positive labels are all the characters belonging to erroneous spans. Furthermore, each true positive is downweighted to half if the system failed to classify the error span's severity (e.g., \textsc{minor} instead of \textsc{major}).
The submitted systems were independently evaluated on in-domain and zero-shot LPs for direct assessments and MQM.  

\subsection{Quality Estimation}

%

\paragraph{Sentence-level.} Results for sentence-level are presented in Table~\ref{tab:results_task_1_DA}. Results indicate that retraining the system from the previous year, specifically \textsc{CometKiwi} with InfoXLM, using data that encompasses this year's DA, leads to significant improvements. Remarkably, this improvement in correlations is achieved while maintaining the same level of correlations for \textit{en-de} (a high-resource language pair for which both models share the same data) and \textit{he-en}, a language pair that both models had not seen during training. Surprisingly, there was a drop in correlations for \textit{zh-en} even though both models saw the same \textit{zh-en} data. Nevertheless, the overall performance of the newly retrained version improved by 4.1 Spearman points.

As anticipated, among the three backbone transformers, the XXL model is the top performer, with significant improvements across all language pairs when compared to InfoXLM. Moreover, additional finetuning on this year's training data results in further improvements for the Indian languages. Notably, concerning the MQM data, this supplementary finetuning step not only preserves performance but sometimes even increases it. Similar to last year, the ensemble of high-performing models once again makes up our best submission.

%
Finally, despite performing well in Task 2, \textsc{GPT4-QE} shows poor correlations at  sentence-level prediction with the exception of the \textit{en-de} for which \textsc{GPT4-QE}, although lagging behind the ensemble approach, surpasses our individual models.

\paragraph{Word-level.} We report the best individual systems Table \ref{tab:results_task_1_word_level_da}. Our best individual systems were trained on top of the InfoXLM L generic model. For PE models, we used multi-task objective in Eq.~\ref{eq:comb_loss}, as we found that combining the sentence-level and word-level loss was beneficial. However, for MQM models, we trained word-level only models, by setting $\lambda_{\textsc{sl}} = 0.0 $ and $\lambda_{\textsc{wl}} = 1.0$.

Interestingly, we found that PE models are very competitive on MQM language pairs. For example, the best overall performance for \textit{he-en} was actually obtained with a PE word-level model. This is also reflected on the Optuna weights obtained for our final ensemble, wherein the weights of the PE models are significantly higher than those of the MQM models for all language pairs but \textit{en-de}. In fact, our final ensemble for \textit{en-zh} and \textit{en-he} consists solely of PE models trained with different learning rates, $\lambda_{\textsc{sl}}$, $\lambda_{\textsc{wl}}$ and $w$. 
Further investigation on two different vectors may lead to improved word-level models: (i)~balancing DA and MQM word-level annotations, and (ii)~appropriately leveraging the larger capacity of scaled up encoder models.  

\paragraph{Fine-grained error span detection.} Results for fine-grained error span detection are shown in Table~\ref{tab:results_task_2}.
Using a word-level model to obtain error span predictions leads to reasonable performance, comparable to our unconstrained submission, \textsc{UnbabelQi}, a model directly tasked with error span detection. That said, x\textsc{comet-ps-ref}, an error span detection model, surpassed both of the previous approaches. We attribute the improved performance to this system being an ensemble of two significantly larger models, and to the usage of a pseudo-reference. We found the latter to be particularly beneficial on \textit{he-en}, a language pair for which we had no training data.
%

The best approach in terms of average $F_1$ was \textsc{GPT4-QE}, mostly due to the improved performance on \textit{en-de}. While this is a promising finding for LLM-based quality estimation systems, there are limitations. First, obtaining a sentence-level score from the error spans (as per the MQM framework) leads to poor correlations with human judgements derived from DA (see Table~\ref{tab:results_task_1_DA}) and with low-resource language-pairs like \textit{he-en}.
%
%
%
Second, despite being useful in practice and leading to gains in $F_1$, it is hard to control GPT's precision and recall. We found that the number of examples included in the prompt, their ordering, and the number of errors within each example led to noticeable changes in the system's propensity to flag errors. Thirdly, running QE with a system such as GPT-4 is expensive and slow even for a shared task exercise.

\begin{table}[t]
    \centering
    \small    
    \begin{tabular}{l ccc c@{} cccc}
        \toprule
        \bf Method & \bf en-de & \bf zh-en & \bf he-en$^\dagger$  & & \bf avg. \\
        \midrule
        Baseline & 0.167 & 0.219 & 0.083 & & 0.156 \\
        2nd place (TBA)& & & & & 0.165 \\
        \midrule
        \textsc{word-level} & 0.235 & \bf 0.272 & 0.105 & & 0.204 \\
        x\textsc{comet-ps-ref} & 0.259 & 0.270 & \bf 0.125 & & 0.218 \\ \cdashlinelr{1-8}
        \textsc{UnbabelQi} & 0.249 & 0.227 & 0.111 & & 0.196 \\
        \textsc{GPT4-QE} & \bf 0.273 &	0.265 & 0.121 & & \bf 0.220 \\
        \bottomrule
    \end{tabular}
    \caption{Results for fine-grained error span detection (Task 2). Evaluation metric is $F_1$ score. We represent zero-shot LPs with $\dagger$. The first two systems are constrained while the other two are unconstrained submissions.
    }
    \label{tab:results_task_2}
\end{table}

\section{Final Remarks}


We describe Unbabel and IST joint submission to WMT23 QE shared task. Our approaches correlate well with human judgements for all the three granularities of translation quality prediction, ranking first in all multilingual tasks and surpassing the previous state-of-the-art model, \textsc{CometKiwi-22}, by up to 10 Spearman correlation points. Overall, our models follow the same architecture of last year's participation, \textsc{CometKiwi}. However, this year we leverage more data and larger encoder models. Our best final systems are ensembles of different models trained on DA, post-edits or MQM scores that complement each other. Interestingly, our best systems surpass GPT-4 by a large margin for sentence-level translation quality prediction, and they are comparable to GPT-4 at error span detection.

\bibliography{anthology, custom}
\bibliographystyle{acl_natbib}

\end{document}

%% file: figs/architecture.tex
\definecolor{paired-light-blue}{RGB}{198, 219, 239}
\definecolor{paired-dark-blue}{RGB}{49, 130, 188}
\definecolor{paired-light-orange}{RGB}{251, 208, 162}
\definecolor{paired-dark-orange}{RGB}{230, 85, 12}
\definecolor{paired-light-green}{RGB}{199, 233, 193}
\definecolor{paired-dark-green}{RGB}{49, 163, 83}
\definecolor{paired-light-purple}{RGB}{218, 218, 235}
\definecolor{paired-dark-purple}{RGB}{117, 107, 176}
\definecolor{paired-light-gray}{RGB}{217, 217, 217}
\definecolor{paired-dark-gray}{RGB}{99, 99, 99}
\definecolor{paired-light-pink}{RGB}{222, 158, 214}
\definecolor{paired-dark-pink}{RGB}{123, 65, 115}
\definecolor{paired-light-red}{RGB}{231, 150, 156}
\definecolor{paired-dark-red}{RGB}{131, 60, 56}
\definecolor{paired-light-yellow}{RGB}{231, 204, 149}
\definecolor{paired-dark-yellow}{RGB}{141, 109, 49}
\tikzset{%
    layernode/.style = {
        align=center,
        text width=6cm,
        inner sep=0.25cm,
        outer sep=0cm,
        rounded corners=4pt,
        fill=paired-light-gray!30,
        draw=paired-dark-gray!50,
    },
    halfnode/.style = {
        align=center,
        text width=2.5cm,
        inner sep=0.25cm,
        outer sep=0cm,
        rounded corners=4pt,
        fill=paired-light-gray!30,
        draw=paired-dark-gray!50,
    },
    outputnode/.style = {
        align=center,
        text width=2.6cm,
    },
    inputnode/.style = {
        align=center,
        inner sep=0.2cm,
    },
}

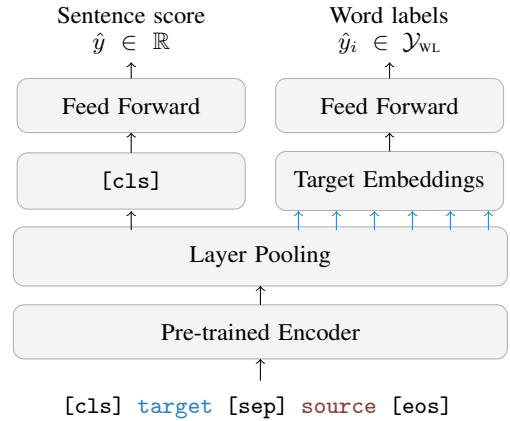
\begin{figure}[t]
    \centering
    \footnotesize	
    \begin{tikzpicture}
        \node[inputnode] (input) at (0, 0) {
            \texttt{
            [cls] \textcolor{paired-dark-blue}{\textbf{target}}  [sep] \textcolor{paired-dark-red}{\textbf{source}} [eos]
            }
        };
        \node[layernode] (encoder) at (0, 1) {Pre-trained Encoder};
        \node[layernode] (scalarmix) at (0, 2) {Layer Pooling};
        \node[halfnode] (avgpool) at (-1.7, 3) {\texttt{[cls]}};
        \node[halfnode] (piece) at (1.7, 3) {Target Embeddings};
        \node[halfnode] (ffn1) at (-1.7, 4) {Feed Forward};
        \node[halfnode] (ffn2) at (1.7, 4) {Feed Forward};
        \node[outputnode] (out1) at (-1.7, 5) {Sentence score \\ $\hat{y} \in \mathbb{R}$};
        \node[outputnode] (out2) at (1.7, 5) {Word labels \\ $\hat{y}_i \in \mathcal{Y}_{\textsc{wl}}$};
        
        \draw[->] (input) to (encoder);
        \draw[->] (encoder) to (scalarmix);
        \draw[->] (-1.7, 2.35) to (-1.7, 2.6);
        \draw[->,paired-dark-blue] (0.5, 2.35) to (0.5, 2.6);
        \draw[->,paired-dark-blue] (1.0, 2.35) to (1.0, 2.6);
        \draw[->,paired-dark-blue] (1.5, 2.35) to (1.5, 2.6);
        \draw[->,paired-dark-blue] (2.0, 2.35) to (2.0, 2.6);
        \draw[->,paired-dark-blue] (2.5, 2.35) to (2.5, 2.6);
        \draw[->,paired-dark-blue] (3.0, 2.35) to (3.0, 2.6);
        \draw[->] (avgpool) to (ffn1);
        \draw[->] (piece) to (ffn2);
        \draw[->] (ffn1) to (out1);
        \draw[->] (ffn2) to (out2);

    \end{tikzpicture}
    
    \caption{Our model follows \textsc{CometKiwi} for sentence-level (left part) and word-level QE (right part). We represent the output space of the word-level head by $\mathcal{Y}_{\textsc{wl}}$.}
    \label{fig:arch}
\end{figure}

%% file: acl_latex.bbl
\begin{thebibliography}{12}
\expandafter\ifx\csname natexlab\endcsname\relax\def\natexlab#1{#1}\fi

\bibitem[{Akiba et~al.(2019)Akiba, Sano, Yanase, Ohta, and
  Koyama}]{optuna_2019}
Takuya Akiba, Shotaro Sano, Toshihiko Yanase, Takeru Ohta, and Masanori Koyama.
  2019.
\newblock \href {https://optuna.org/} {Optuna: A next-generation hyperparameter
  optimization framework}.
\newblock In \emph{Proceedings of the 25rd {ACM} {SIGKDD} International
  Conference on Knowledge Discovery and Data Mining}.

\bibitem[{Chi et~al.(2021)Chi, Dong, Wei, Yang, Singhal, Wang, Song, Mao,
  Huang, and Zhou}]{chi-etal-2021-infoxlm}
Zewen Chi, Li~Dong, Furu Wei, Nan Yang, Saksham Singhal, Wenhui Wang, Xia Song,
  Xian-Ling Mao, Heyan Huang, and Ming Zhou. 2021.
\newblock \href {https://doi.org/10.18653/v1/2021.naacl-main.280} {{I}nfo{XLM}:
  An information-theoretic framework for cross-lingual language model
  pre-training}.
\newblock In \emph{Proceedings of the 2021 Conference of the North American
  Chapter of the Association for Computational Linguistics: Human Language
  Technologies}, pages 3576--3588, Online. Association for Computational
  Linguistics.

\bibitem[{Conneau et~al.(2020)Conneau, Khandelwal, Goyal, Chaudhary, Wenzek,
  Guzm{\'a}n, Grave, Ott, Zettlemoyer, and
  Stoyanov}]{conneau-etal-2020-unsupervised}
Alexis Conneau, Kartikay Khandelwal, Naman Goyal, Vishrav Chaudhary, Guillaume
  Wenzek, Francisco Guzm{\'a}n, Edouard Grave, Myle Ott, Luke Zettlemoyer, and
  Veselin Stoyanov. 2020.
\newblock \href {https://doi.org/10.18653/v1/2020.acl-main.747} {Unsupervised
  cross-lingual representation learning at scale}.
\newblock In \emph{Proceedings of the 58th Annual Meeting of the Association
  for Computational Linguistics}, pages 8440--8451, Online. Association for
  Computational Linguistics.

\bibitem[{Fernandes et~al.(2023)Fernandes, Deutsch, Finkelstein, Riley,
  Martins, Neubig, Garg, Clark, Freitag, and Firat}]{Fernandes2023AutoMQM}
Patrick Fernandes, Daniel Deutsch, Mara Finkelstein, Parker Riley, Andr{\'e}
  F.~T. Martins, Graham Neubig, Ankush Garg, J.~Clark, Markus Freitag, and
  Orhan Firat. 2023.
\newblock \href {https://api.semanticscholar.org/CorpusID:260886800} {The devil
  is in the errors: Leveraging large language models for fine-grained machine
  translation evaluation}.
\newblock \emph{ArXiv}, abs/2308.07286.

\bibitem[{Fomicheva et~al.(2022)Fomicheva, Sun, Fonseca, Blain, Chaudhary,
  Guzm\'an, Lopatina, Specia, and Martins}]{fomicheva2020mlqepe}
Marina Fomicheva, Shuo Sun, Erick Fonseca, Fr\'ed\'eric Blain, Vishrav
  Chaudhary, Francisco Guzm\'an, Nina Lopatina, Lucia Specia, and Andr\'e F.~T.
  Martins. 2022.
\newblock \href {https://aclanthology.org/2022.lrec-1.530} {{MLQE-PE: A
  Multilingual Quality Estimation and Post-Editing Dataset}}.
\newblock In \emph{Proceedings of the Language Resources and Evaluation
  Conference}, pages 4963--4974, Marseille, France. European Language Resources
  Association.

\bibitem[{Martins and Astudillo(2016)}]{martins2016softmax}
Andre Martins and Ramon Astudillo. 2016.
\newblock \href {http://proceedings.mlr.press/v48/martins16.pdf} {From softmax
  to sparsemax: A sparse model of attention and multi-label classification}.
\newblock In \emph{International Conference on Machine Learning}, pages
  1614--1623.

\bibitem[{OpenAI(2023)}]{openai2023gpt4}
OpenAI. 2023.
\newblock \href {http://arxiv.org/abs/2303.08774} {Gpt-4 technical report}.

\bibitem[{Peters et~al.(2018)Peters, Neumann, Iyyer, Gardner, Clark, Lee, and
  Zettlemoyer}]{peters-etal-2018-deep}
Matthew~E. Peters, Mark Neumann, Mohit Iyyer, Matt Gardner, Christopher Clark,
  Kenton Lee, and Luke Zettlemoyer. 2018.
\newblock \href {https://doi.org/10.18653/v1/N18-1202} {Deep contextualized
  word representations}.
\newblock In \emph{Proceedings of the 2018 Conference of the North {A}merican
  Chapter of the Association for Computational Linguistics: Human Language
  Technologies, Volume 1 (Long Papers)}, pages 2227--2237, New Orleans,
  Louisiana. Association for Computational Linguistics.

\bibitem[{Rei et~al.(2022{\natexlab{a}})Rei, Farinha, de~Souza, Ramos, Martins,
  Coheur, and Lavie}]{rei-etal-2022-searching}
Ricardo Rei, Ana~C Farinha, Jos{\'e}~G.C. de~Souza, Pedro~G. Ramos,
  Andr{\'e}~F.T. Martins, Luisa Coheur, and Alon Lavie. 2022{\natexlab{a}}.
\newblock \href {https://aclanthology.org/2022.eamt-1.9} {Searching for
  {COMETINHO}: The little metric that could}.
\newblock In \emph{Proceedings of the 23rd Annual Conference of the European
  Association for Machine Translation}, pages 61--70, Ghent, Belgium. European
  Association for Machine Translation.

\bibitem[{Rei et~al.(2022{\natexlab{b}})Rei, Treviso, Guerreiro, Zerva,
  Farinha, Maroti, C.~de Souza, Glushkova, Alves, Coheur, Lavie, and
  Martins}]{rei-etal-2022-cometkiwi}
Ricardo Rei, Marcos Treviso, Nuno~M. Guerreiro, Chrysoula Zerva, Ana~C Farinha,
  Christine Maroti, Jos{\'e}~G. C.~de Souza, Taisiya Glushkova, Duarte Alves,
  Luisa Coheur, Alon Lavie, and Andr{\'e} F.~T. Martins. 2022{\natexlab{b}}.
\newblock \href {https://aclanthology.org/2022.wmt-1.60} {{C}omet{K}iwi:
  {IST}-unbabel 2022 submission for the quality estimation shared task}.
\newblock In \emph{Proceedings of the Seventh Conference on Machine Translation
  (WMT)}, pages 634--645, Abu Dhabi, United Arab Emirates (Hybrid). Association
  for Computational Linguistics.

\bibitem[{Specia et~al.(2018)Specia, Scarton, and Paetzold}]{specia2018quality}
Lucia Specia, Carolina Scarton, and Gustavo~Henrique Paetzold. 2018.
\newblock \href {https://ieeexplore.ieee.org/document/8473393} {{Quality
  estimation for machine translation}}.
\newblock \emph{Synthesis Lectures on Human Language Technologies},
  11(1):1--162.

\bibitem[{Wan et~al.(2022)Wan, Liu, Yang, Zhang, Chen, Wong, and
  Chao}]{wan-etal-2022-unite}
Yu~Wan, Dayiheng Liu, Baosong Yang, Haibo Zhang, Boxing Chen, Derek Wong, and
  Lidia Chao. 2022.
\newblock \href {https://doi.org/10.18653/v1/2022.acl-long.558} {{U}ni{TE}:
  Unified translation evaluation}.
\newblock In \emph{Proceedings of the 60th Annual Meeting of the Association
  for Computational Linguistics (Volume 1: Long Papers)}, pages 8117--8127,
  Dublin, Ireland. Association for Computational Linguistics.

\end{thebibliography}
